\documentclass[runningheads]{llncs}

\usepackage{eccv}

\usepackage{multirow}
\usepackage{footmisc}
\usepackage{wrapfig}

\usepackage{eccvabbrv}

\usepackage{graphicx}
\usepackage{booktabs}

\usepackage[accsupp]{axessibility}  

\usepackage{hyperref}

\usepackage{orcidlink}

\begin{document}

\title{Temporally Consistent Stereo Matching} 

\titlerunning{Temporally Consistent Stereo Matching}

\author{Jiaxi Zeng\inst{1,2}\orcidlink{0009-0007-0481-2005} \and
Chengtang Yao\inst{1,3} \and
Yuwei Wu\inst{1,2}\thanks{Corresponding author} \and
Yunde Jia\inst{2,1}}

\authorrunning{J. Zeng et al.}

\institute{Beijing Key Laboratory of Intelligent Information Technology, \\ School of Computer Science \& Technology, Beijing Institute of Technology, China \and
Guangdong Laboratory of Machine Perception and Intelligent Computing, \\
Shenzhen MSU-BIT University, China \and
Horizon Robotics \\
\email{\{jiaxi,yao.c.t,wuyuwei,jiayunde\}@bit.edu.cn}}

\maketitle

\begin{abstract}
Stereo matching provides depth estimation from binocular images for downstream applications. These applications mostly take video streams as input and require temporally consistent depth maps. However, existing methods mainly focus on the estimation at the single-frame level. This commonly leads to temporally inconsistent results, especially in ill-posed regions.
In this paper, we aim to leverage temporal information to improve the temporal consistency, accuracy, and efficiency of stereo matching. To achieve this, we formulate video stereo matching as a process of temporal disparity completion followed by continuous iterative refinements. 
Specifically, we first project the disparity of the previous timestamp to the current viewpoint, obtaining a semi-dense disparity map. Then, we complete this map through a disparity completion module to obtain a well-initialized disparity map. 
The state features from the current completion module and from the past refinement are fused together, providing a temporally coherent state for subsequent refinement. 
Based on this coherent state, we introduce a dual-space refinement module to iteratively refine the initialized result in both disparity and disparity gradient spaces, improving estimations in ill-posed regions.
Extensive experiments demonstrate that our method effectively alleviates temporal inconsistency while enhancing both accuracy and efficiency. Currently, our method ranks second on the KITTI 2015 benchmark, while achieving superior efficiency compared to other state-of-the-art methods. The code is available at \url{https://github.com/jiaxiZeng/Temporally-Consistent-Stereo-Matching}.
\end{abstract}

\section{Introduction}
\label{sec:intro}
\begin{figure*}[!t]
\centering
\includegraphics[width=0.9\linewidth]{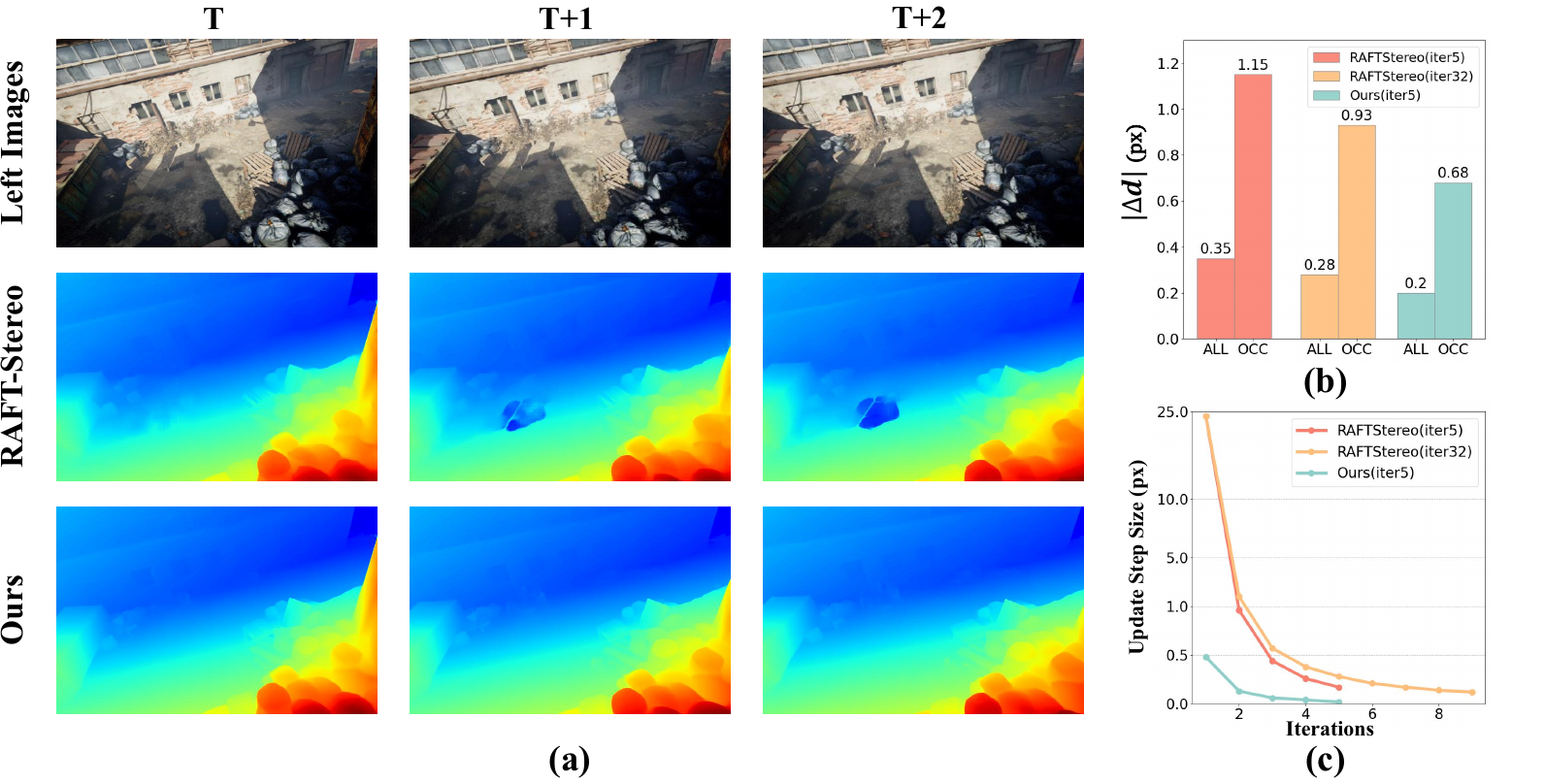}
\caption{(a) Visualization of disparity sequences from RAFT-Stereo\cite{lipson2021raft} and our TC-Stereo. The failure cases of RAFT-Stereo lie in the reflective areas on the ground.
A small motion can cause severe jitters in disparity predictions, while our method achieves temporally stable outputs. (b) The jitter $|\Delta d|$ between aligned successive disparity maps in all and occluded areas on TartanAir dataset\cite{wang2020tartanair}. Our method achieves better temporal consistency than RAFT-Stereo. (c) The update step size changed with the iterations of RAFT-Stereo and our TC-Stereo on TartanAir. Compared to RAFT-Stereo, our method performs a disparity search within a local range.}
\label{fig:1}
\end{figure*}

Stereo matching estimates the disparity by finding the horizontal correspondences between the rectified left and right images.
It plays a significant role in various fields such as autonomous driving\cite{menze2015object,yang2019drivingstereo}, robotics\cite{usenko2016direct}, SLAM\cite{murORB2, campos2021orb}, AR/VR\cite{wang2023practical,cheng2024stereo}.  
In these applications, stereo videos are typically taken as input and the output disparity sequences are employed for downstream tasks. 
However, the majority of existing methods\cite{lipson2021raft,li2022practical,xu2023iterative,shen2021cfnet,xu2022acvnet} predict disparities independently for each stereo pair, disregarding the coherence between consecutive frames.
This typically results in temporally inconsistent results, as shown in the second row of Fig. \ref{fig:1} (a). The inconsistent depth information has been demonstrated to substantially influence the accuracy and consistency of downstream tasks\cite{wang2022neuris,zhang2023go,deng2022depth}. Although some techniques, such as Extended Kalman Filtering (EKF)\cite{kalman1960new} or Bundle Adjustment (BA)\cite{triggs2000bundle}, mitigate the impact of inconsistency, they also have respective limitations. For instance, EKF suffers from non-linear errors, while BA brings intensive computation.

In this work, our goal is to improve the temporal consistency of stereo matching.
We analyze the source of temporal inconsistency in stereo matching from two perspectives.
On the one hand, most of the existing methods \cite{chang2018pyramid,shen2021cfnet,lipson2021raft} independently infer the disparity map from scratch for each frame. 
This leads to a global disparity searching range, which introduces greater variability in disparity computation (\eg, larger update step size), increasing the likelihood of inconsistency between consecutive frames.
As shown in Fig. \ref{fig:1} (c), our method has a smaller update step size, indicating that it searches for the ground truth within a local disparity range, while RAFT-Stereo regresses the disparity from scratch in a global disparity range.
On the other hand, camera or object motion leads to appearance changes in occluded areas, reflective surfaces, or low-texture regions.  
The inherent ambiguity in these ill-posed regions causes the model to output unstable results when processing the temporally changing image sequences. Fig. \ref{fig:1} (b) demonstrates that the temporal inconsistency in ill-posed regions, like occlusions, is more serious than that in general regions.

Based on the above analyses, we propose to leverage the temporal information to mitigate temporal inconsistency. To avoid large update steps during the refinement process, we use the semi-dense disparity map from the previous time step for disparity completion, providing a well-initialized disparity for subsequent iterative refinement. This allows the refinement module to focus on local-range disparity updates. Additionally, we conduct a straightforward temporal state fusion to fuse the hidden state from the previous refinement with the current state features from the completion module, thus providing a temporally coherent initial hidden state for further refinement.

Moreover, we find that, for ill-posed areas like occlusions or reflective regions, it is easier to estimate the disparity gradient than the disparity itself. This is primarily because, in the real world, the depth of most regions tends to be smooth and continuous, producing constant or gradually changing gradients in these areas. Based on this observation, we propose a dual-space refinement module. This module takes the temporally initialized disparity and fused states as inputs, and refines the results in both disparity space and disparity gradient space. Through iterative refinement, the local smoothness constraints in the disparity gradient space are progressively extended to more global areas, thereby improving the smoothness of ill-posed regions and resulting in stable outputs.

Extensive experiments show that our method achieves state-of-the-art temporal consistency and accuracy. 
The contributions of our work can be summarized as follows: 
(1) We analyze the causes of temporal inconsistency and propose a temporally consistent stereo matching method. 
(2) We propose a temporal disparity completion and a temporal state fusion module to exploit temporal information, providing a well-initialized disparity and a coherent hidden state for refinement. 
(3) We propose an iterative dual-space refinement module to refine and stabilize the results in ill-posed regions. 
(4) Our method improves the temporal consistency of stereo matching and achieves SOTA results on synthetic and real-world datasets.

\section{Related Work}
\noindent\textbf{Deep Stereo Matching} Existing deep stereo matching methods primarily revolve around cost volume to design networks or representations, which is commonly categorized into regression-based methods\cite{kendall2017end,chang2018pyramid,guo2019group,khamis2018stereonet,xu2020aanet,duggal2019deeppruner,gu2020cascade,shen2021cfnet,xu2022acvnet} and iterative-based methods\cite{lipson2021raft,li2022practical,xu2023iterative,zeng2023parameterized}. Regression-based methods regress a probability volume to compute the disparity map, which can be classified into 3D volume methods\cite{mayer2016large,liang2017learning,tonioni2019real,xu2020aanet} and 4D volume methods\cite{kendall2017end,chang2018pyramid,duggal2019deeppruner,shen2021cfnet,mao2021uasnet,xu2022acvnet}. They either directly regress the disparity in the global disparity range \cite{kendall2017end,chang2018pyramid,xu2022acvnet} or perform a global-to-local regression \cite{duggal2019deeppruner,shen2021cfnet,mao2021uasnet}. 
However, the motion between frames may cause variation in the global disparity probability distribution, leading to inconsistencies between sequential disparity maps.
Iterative-based methods regard stereo matching as a continuous optimization process in the disparity space, iteratively retrieving the cost volume to refine the disparity map. 
RAFT-Stereo\cite{lipson2021raft} leverages the optical flow method RAFT \cite{teed2020raft} for stereo matching and introduces a multi-level GRU to enlarge the receptive field. CREStereo\cite{li2022practical} introduces a cascade network and adaptive group correlation layer to address challenges such as thin structures and imperfect rectification. DLNR\cite{zhao2023high} employs LSTM to alleviate the data coupling problem in the iteration of GRU. 
PCVNet\cite{zeng2023parameterized} proposed a parameterized cost volume to accelerate the convergence of iterations by predicting larger update steps. 
IGEV-Stereo\cite{xu2023iterative} regresses a geometry encoding volume to provide an accurate initial disparity for the GRU module, accelerating the convergence of disparity optimization. 
Although these methods have achieved excellent results, independently inferring the disparities for each frame leads to poor temporal consistency. In contrast, our method integrates the disparity and state information of the previous frame to ensure coherence between the consecutive outputs.

\noindent\textbf{Video Stereo Matching} 
Recently, there has been increasing attention on video stereo matching. Dynamic-Stereo\cite{karaev2023dynamicstereo} and CODD\cite{li2023temporally} focus on consistent stereo matching in dynamic scenes. Dynamic-Stereo jointly processes multiple frames, improving the temporal consistency of the results. However, the latency and computational overhead introduced by multi-frame inference limit its application in online scenarios.
CODD predicts SE(3) transformations for each pixel to align successive disparity maps and fuses them together. Different from their post-fusion strategy, we integrate temporal information before our refinement, providing a robust local disparity searching range.
TemporalStereo \cite{zhang2023temporalstereo} mitigates the adverse effects of occlusions and reflections by augmenting current cost volume with the costs from the previous frame. Nevertheless, it still relies on coarse-to-fine regression in the global disparity range.
XR-Stereo\cite{cheng2024stereo} extends RAFT-Stereo\cite{lipson2021raft} over time, reducing iterations by reusing results from the previous frame, achieving over 100 FPS in a low-resolution mode. However, it simply takes the previous disparity and hidden state as the initialization of the current frame, disregarding the initial disparity and hidden state are semi-dense due to non-overlapping regions between frames. For these regions, it is still necessary to regress the disparity from scratch. Instead, we suggest completing the disparity map and incorporating the current state information to obtain a robust initialization for local refinement. The importance of these designs for improving temporal consistency was demonstrated in our ablation experiments.

\noindent\textbf{Depth Completion} Depth completion relies on a sparse depth map to generate a dense depth map\cite{ma2018sparse,tang2020learning,cheng2018depth,cheng2020cspn++}. Some methods\cite{bartolomei2023revisiting,choi2021stereo} formulate depth completion as a stereo matching task, while some stereo matching methods\cite{bartolomei2023active,wang20193d} utilizing sparse depth from LiDAR to guide the matching process.
Different from these works, our disparity completion module does not rely on additional LiDAR data or complicated architectures. It utilizes a simple encoder-decoder to complete a semi-dense disparity map projected from the previous time stamp and provide the state features of the completed disparity map for further refinement.

\noindent\textbf{Normal Guided Depth Estimation}
Substantial research \cite{qi2018geonet,yin2019enforcing,long2021adaptive,tankovich2021hitnet,long2020occlusion,kusupati2020normal} demonstrates that utilizing surface normal contributes to depth estimation. Most methods \cite{yin2019enforcing,long2021adaptive,long2020occlusion,kusupati2020normal} design loss functions to leverage normal for constraining the depth map.
Geo-Net\cite{qi2018geonet} utilizes constraints between the normal and depth during inference, but this constraint is limited to local areas.
HITNet\cite{tankovich2021hitnet} up-samples the disparity map through predicted disparity gradients to constrain the local surfaces. In this work, we iteratively refine results in the disparity and disparity gradient spaces, propagating the surface constraint globally.

\section{Method}
\begin{figure*}[!t]
\centering
\includegraphics[width=0.95\linewidth]{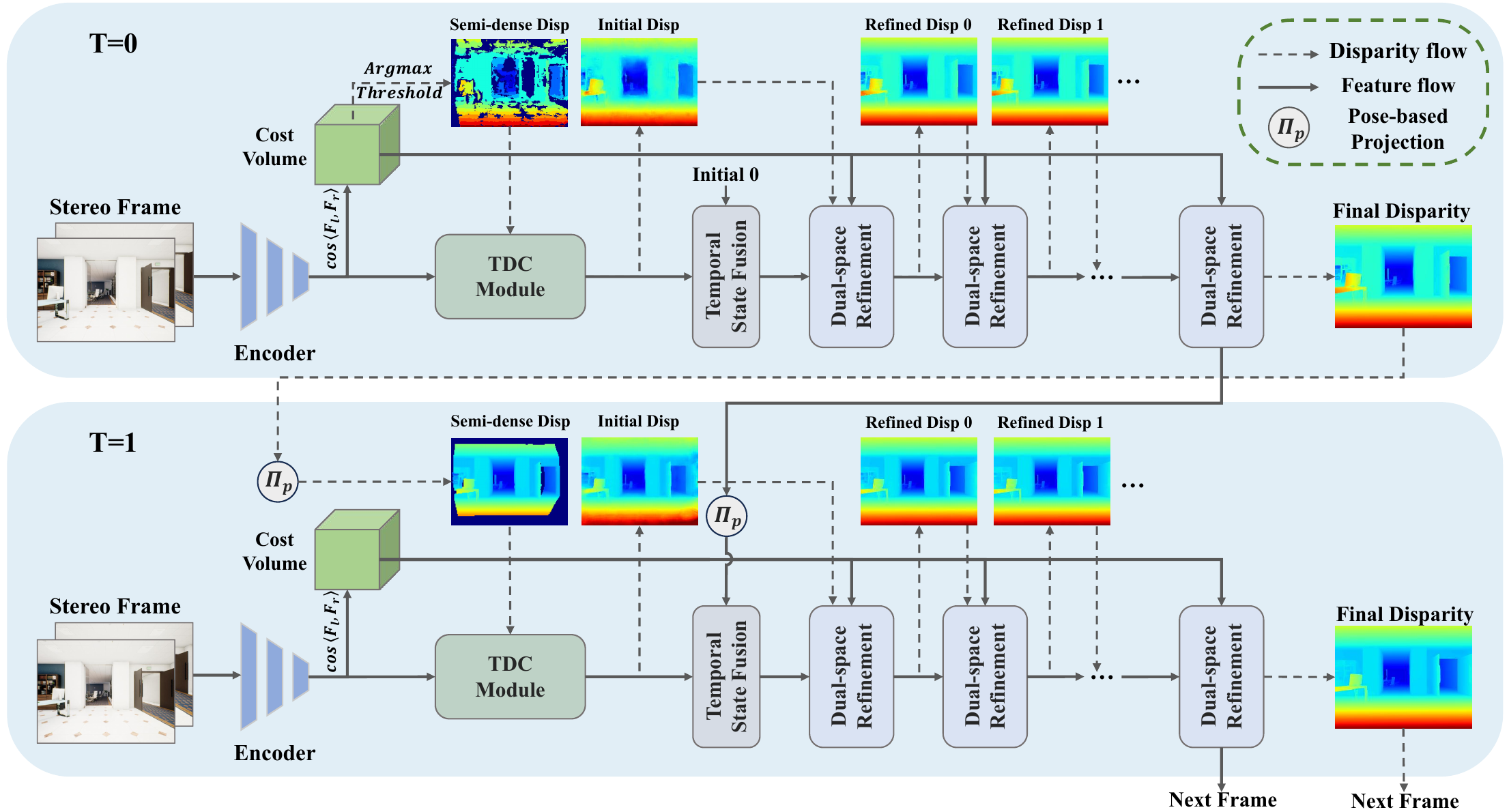}
\caption{\textbf{Pipeline of TC-Stereo.} We first use an encoder to extract left and right features for the current stereo frame. These features are then used to construct a cost volume. A semi-dense disparity map, derived from the cost volume (for the first frame) or projected from the previous timestamp (for subsequent frames), is fed into the Temporal Disparity Completion (TDC) module to obtain an initial dense disparity map. The output state of the TDC module is fused with the state from the past to provide an initial hidden state for refinement. The dual-space refinement module iteratively retrieves the cost volume and alternately refines the disparity map in the disparity and disparity gradient spaces. The final disparity map and hidden state are projected into the viewpoint of the next frame, serving as the temporal information for continuous disparity estimation.
}
\label{fig:pipeline}
\end{figure*}

Our method processes stereo video sequences in an online manner and outputs the temporally consistent disparity maps. The pipeline of TC-Stereo is depicted in Fig. \ref{fig:pipeline}. 
In the following subsections, we will provide a detailed introduction to three key components of our TC-Stereo: temporal disparity completion (Sec. \ref{sec:completion}), temporal state fusion (Sec. \ref{sec:state fusion}), and iterative dual-space refinement (Sec. \ref{sec:refinement}). Finally, we will present the loss functions involved (Sec. \ref{loss}).

\subsection{Temporal Disparity Completion\label{sec:completion}}
The aim of temporal disparity completion is to provide a well-initialized disparity map by leveraging the result of the previous frame.
However, for the initial frame, temporal information is unattainable. As a result, there is no disparity point to use as the hint for the completion module, causing it to degrade into a monocular module and affecting the reliability of the initialization.
To address this, we propose leveraging a semi-dense disparity map derived from the cost volume to compensate for the lack of temporal information in the initial frame.

\noindent\textbf{Semi-dense disparity map from the cost volume}  We define the cost volume \( C \in [0,1]^{H \times W \times D}\) as the cosine similarity between the left and right features for each disparity hypothesis \( d \in \mathbb D=\{0,1,...,D-1\} \). This is expressed as:
{
\begin{equation}
\begin{aligned}
C(v,u,d) = \frac{\langle F_l(v,u), F_r(v,u-d) \rangle}{\|F_l(v,u)\|_2 \times \|F_r(v,u-d)\|_2},
\end{aligned}
\label{Eq: cos sim}
\end{equation}}

\noindent where \( F_l(v,u) \) and \( F_r(v,u-d) \) represent the feature vectors at pixel coordinates \( (v,u) \) in the left image and \( (v,u-d) \) in the right image, respectively. Here, $\langle\cdot,\cdot\rangle$ denotes the inner product and $\|\cdot\|_2$ is Euclidean norm of a feature vector.
We first compute the disparity from the cost volume by the winner-take-all strategy. Then, a threshold is applied to filter out outliers and obtain a semi-dense disparity map. This process can be summarized by the following formula:
{
\begin{equation}
\begin{aligned}
&d_{\text{1}} = {\arg\max}_{d \in \mathbb D}  C(d), \\
&d_{\text{2}} = {\arg\max}_{d \in \mathbb D \setminus \{d_{\text{1}}-1,d_{\text{1}},d_{\text{1}}+1\}}  C(d), \\
&d{_\text{s}} = \begin{cases} 
d_{\text{1}}, & \text{if } C(d_{\text{1}}) - C(d_{\text{2}}) > \theta \\
0, & \text{otherwise}
\end{cases},
\end{aligned}
\label{Eq: argmax with threshold}
\end{equation}}

\noindent where $d_1$ is the disparity with the highest similarity, and $d_2$ is the second-highest disparity, excluding the immediate neighbors of $d_1$. The semi-dense disparity map $d_{\text{s}}$ retains $d_1$ only if the $C(d_1)$ exceeds the $C(d_2)$ by a threshold $\theta$; otherwise, it is set to 0, indicating an invalid point. This process filters out disparities with high uncertainty, yielding a more reliable semi-dense disparity map for the completion. 
Moreover, to effectively learn the semi-dense disparity map from the cost volume, we apply a contrastive loss\cite{tankovich2021hitnet} to the cost volume. We discuss the loss in detail in Section \ref{loss}.

\noindent\textbf{Semi-dense disparity map from the previous timestamp} 
For subsequent frames, we project the previous disparity map into the current image coordinates based on the intrinsic parameters and poses. This process can be represented as:
{
\begin{equation}
\centering
\begin{aligned}
&p^{t-1 \rightarrow t} = \Pi_c(T^{t-1 \rightarrow t} \cdot \Pi_c^{-1}(p^{t-1},z^{t-1})), \\
&z^{t-1 \rightarrow t} = (T^{t-1 \rightarrow t} \cdot \Pi_c^{-1}(p^{t-1},z^{t-1}))_z, \\
&d^{t-1 \rightarrow t} = \frac{bf}{z^{t-1 \rightarrow t}}, \\
&d^{t}_{\text{s}} = \text{warp}(d^{t-1 \rightarrow t},p^{t-1 \rightarrow t}).
\end{aligned}
\label{projection}
\end{equation}}

\noindent Here, \( p^{t-1} \) and \( z^{t-1} \) denote the coordinates of pixels and depths at time \( t-1 \), respectively. \(\Pi_c\) represents the image projection process, which maps 3D points onto the 2D image.
\(T^{t-1 \rightarrow t}\) is the relative transformation from time $t-1$ to $t$. $b$ is the baseline of stereo camera, and $f$ is the pixel-presented focal length. We forward warp $d^{t-1 \rightarrow t}$ by the pixel mapping $p^{t-1 \rightarrow t}$ to get the semi-dense disparity map $d_s^{t}$ at time $t$. 
The flexible sources of the semi-dense disparity map enable our model to perform inference in both single-frame and multi-frame modes.

\noindent\textbf{Disparity completion} Our temporal disparity completion module utilizes contexts from the feature encoder, the semi-dense disparity, and a binary mask that indicates the sparsity as inputs. It employs a lightweight encoder-decoder network to regress a dense disparity map and outputs state features of the map for temporal state fusion. The detailed architecture of the completion module is provided in the supplementary materials.

\subsection{Temporal State Fusion\label{sec:state fusion}}
Different from XR-Stereo\cite{cheng2024stereo}, which directly uses the warped previous hidden states to initialize the refinement module, we argue that the previous states may not be the best choice for initializing the current states.
On the one hand, the past states only encode the state information along the previous line of sight, which may fail in the current viewpoint.
On the other hand, for non-overlapping regions between two frames, there are no past hidden states available.
Therefore, we leverage a lightweight GRU-like module to fuse the current state features $c^{t}$ from the TDC module with the past hidden states $h^{t-1}_{N}$ (N denotes the number of
refinement iterations) to provide initial hidden states $h^{t}_0$ for the current refinement module:
{
\begin{equation}
    \begin{aligned}
        z^t &= \sigma(W_z \cdot [c^t, h^{t-1}_{N-1}]), \\
        r^t &= \sigma(W_r \cdot [c^t, h^{t-1}_{N-1}]), \\
        q^t &= \tanh(W_q \cdot [r^t \odot c^t, h^{t-1}_{N-1}]), \\
        h^t_0 &= z^t \odot c^t + (1-z^t) \odot q^t .
    \end{aligned}
\end{equation}
}

\noindent Here, $W_z$, $W_r$, and $W_q$ are the parameters of the network, $\sigma(\cdot)$ denotes the sigmoid function, and $\odot$ is the element-wise dot. This simple module plays a crucial role in enhancing stability and accuracy, which we will demonstrate in the experimental section.

\begin{figure*}[!t]
\centering
\includegraphics[width=0.98\linewidth]{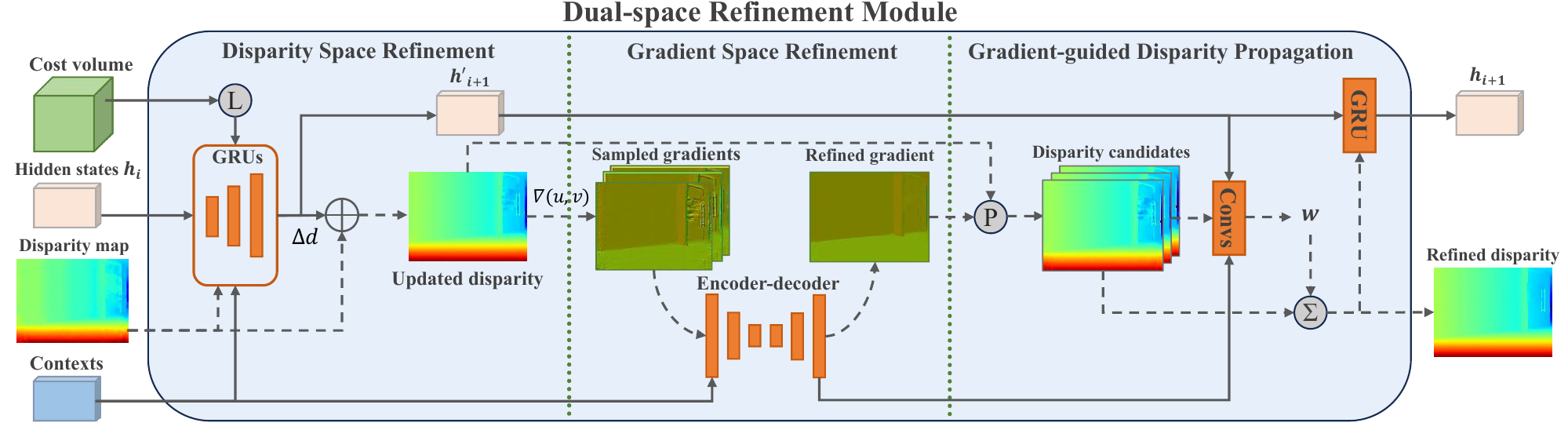}
\caption{\textbf{Architecture of the dual-space refinement module.} The disparity map corresponds to a scene consisting of a wall, a floor, and a transparent glass door.
The encircled \textbf{L} denotes the lookup operation to the cost volume, \textbf{P} represents the gradient-guided disparity propagation, and $\mathbf \Sigma$ means weighted summation.
}
\label{fig:fig3}
\end{figure*}

\subsection{Dual-space Refinement\label{sec:refinement}}
We observed that for some challenging regions for the disparity space, such as reflective areas, the gradient values can be easily estimated in the disparity gradient space. Based on this observation, we design a dual-space refinement module to refine the results in both the disparity space and the gradient space. As shown in Fig. \ref{fig:fig3}, the pipeline of the dual-space module consists of three steps: disparity space refinement, gradient space refinement, and gradient-guided disparity propagation.

\noindent\textbf{Disparity space refinement} Follow RAFT-Stereo\cite{lipson2021raft}, we leverage the multi-level GRUs to refine the disparity map. It takes the hidden state $h_i$ ($i$ denotes the iteration), contextual features, the looked-up costs, and the disparity map as inputs. It outputs an intermediate hidden state $h'_{i+1}$ and a step size $\Delta d$ to update the disparity map. 

\noindent\textbf{Gradient space refinement} We convert the updated disparity map to its disparity gradient space. For a point $(u_0, v_0, d_0)$ in the disparity space, we sample two neighbor points $(u_1, v_1, d_1)$ and $(u_2, v_2, d_2)$, which yields two vectors, $\mathbf{x}_1=(\Delta u_1,\Delta v_1, \Delta d_1)=(u_1-u_0,v_1-v_0,d_1-d_0)$ and $\mathbf{x}_2=(\Delta u_2,\Delta v_2, \Delta d_2)=(u_2-u_0,v_2-v_0,d_2-d_0)$. We assume that \( \mathbf{x}_1 \) and \( \mathbf{x}_2 \) are non-collinear, and the rotation from \( \mathbf{x}_1 \) to \( \mathbf{x}_2 \) in the $u$-$v$ coordinate is clockwise. The formula of the disparity gradient can be represented as:
\begin{equation}
\renewcommand\arraystretch{1.5}
\begin{array}{c}
\begin{pmatrix}
\frac{\partial d}{\partial u} \\
\frac{\partial d}{\partial v}
\end{pmatrix}
= 
\begin{pmatrix} 
\frac{\Delta v_1 \Delta d_2 - \Delta v_2 \Delta d_1}{\Delta u_2 \Delta v_1 - \Delta u_1 \Delta v_2} \\
\frac{\Delta u_2 \Delta d_1 - \Delta u_1 \Delta d_2}{\Delta u_2 \Delta v_1 - \Delta u_1 \Delta v_2} 
\end{pmatrix}
\end{array}.
\end{equation}
By sampling different points within the neighborhood, we can obtain a series of gradient maps. The sampled gradients and contextual features are input into an encoder-decoder network to regress a refined gradient map.

\noindent\textbf{Gradient-guided disparity propagation} We utilize the refined gradients to guide the propagation of refined disparity in the disparity space. Specifically, for each pixel $(u,v)$ with disparity $d$, we propagate the disparity to its neighborhood $(u_n,v_n)$ based on the local planar hypothesis:
{\begin{equation}
\hat{d}_n=d+(u_n-u)\frac{\partial d}{\partial u}+(v_n-v)\frac{\partial d}{\partial v},
\end{equation}}

\noindent where $\hat{d}_n$ is a propagated disparity candidate. We concatenated the $h'_{i+1}$, contexts, and the propagated disparity candidates and fed them into several convolutional layers with a softmax function to regress a set of weights, denoted as $w$. The refined disparity is regressed through the weighted summation of the disparity candidates. Finally, a lightweight GRU network with 1×1 convolutional layers updates the intermediate hidden state using the refined disparity. The updated hidden state $h_{i+1}$ and the disparity are then used as inputs for the next iteration.
By iteratively refining the results in both disparity and gradient space, the surface constraints are propagated globally, improving results in ill-posed areas.  

\subsection{Loss functions\label{loss}}
Our loss function consists of three components: the cost volume loss \(\mathcal{L}_{cv}\), the disparity loss \(\mathcal{L}_{disp}\), and the disparity gradient loss \(\mathcal{L}_{grad}\). 
For the \(\mathcal{L}_{cv}\), we adopt the contrastive loss proposed by HITNet\cite{tankovich2021hitnet} to supervise the cost volume:
{
\begin{equation}
\begin{aligned}
&\mathcal{L}_{cv} = 1 - \psi(d^{gt}) + \max(\eta + \psi(d^{nm}) - \psi_{detach}(d^{gt}), 0), \\
&d^{nm} = {\arg\max}_{d \in \mathbb{D} \setminus [d^{gt} - 1.5, d^{gt} + 1.5]} C(d), \\
&\psi(d) = (d - \lfloor d \rfloor) C(\lfloor d \rfloor + 1) + (\lfloor d \rfloor + 1 - d) C(\lfloor d \rfloor).
\end{aligned}
\end{equation}}


\noindent $\psi(d)$ is the cost (strictly speaking, similarity) for a sub-pixel disparity $d$. It is obtained through linear interpolation between $C(\lfloor d \rfloor + 1)$ and $C(\lfloor d \rfloor)$, where $\lfloor\cdot\rfloor$ is the floor function. The aim of \(\mathcal{L}_{cv}\) is to maximize \(\psi(d^{gt})\), while penalizing \(\psi(d^{nm})\) to ensure it remains at least a threshold \(\eta\) lower than \(\psi(d^{gt})\).

As for the disparity loss $\mathcal{L}_{disp}$, it comprises three subparts: \textbf{d}isparity \textbf{c}ompletion loss $\mathcal{L}_{dc}$, \textbf{d}isparity \textbf{s}pace \textbf{r}efinement loss $\mathcal{L}_{dsr}$, and \textbf{g}radient-guided \textbf{d}isparity \textbf{p}ropagation loss $\mathcal{L}_{gdp}$, which can be expressed as: 
{ \begin{equation}
\begin{aligned}
\mathcal{L}_{disp} = \lambda_{dc}\mathcal{L}_{dc} + \sum_{i=1}^N\gamma^{N-i}(\mathcal{L}^i_{dsr} + \lambda_{gdp}\mathcal{L}^i_{gdp}).
\end{aligned}
\end{equation}}

\noindent \( N \) represents the number of iterations, \( \gamma \) denotes the decay coefficient, and \( \lambda_{dc} \) and \( \lambda_{gdp}\)  represent the balancing scalars.
All these loss functions utilize the L1 loss between the $d^{gt}$ and the disparity outputs at the corresponding stage.

The disparity gradient losses $\mathcal{L}_{grad}$ can be formulated as:
{ \begin{equation}
\begin{aligned}
\mathcal{L}_{grad} =\|&g^{gsr}-g^{gt}\|_1 +\|g^{gdp}-g^{gt}\|_1, \\
g^{gt}&=\nabla_{u,v} d^{gt}, \\
g^{gdp}&=\nabla_{u,v} d^{gdp}.
\end{aligned}
\end{equation}}

\noindent The \(g^{gsr}\) represents the refined gradient map in the \textbf{g}radient \textbf{s}pace \textbf{r}efinement. The $g^{gt}$ and $g^{gdp}$ are derived by taking the gradients of $d^{gt}$ and $d^{gdp}$ with respect to \(u\) and \(v\).
The final loss is the combination of $\mathcal{L}_{cv}$, $\mathcal{L}_{disp}$ and $\mathcal{L}_{grad}$:
{ \begin{equation}
\begin{aligned}
\mathcal{L} = \mathcal{L}_{cv}+\mathcal{L}_{disp}+\mathcal{L}_{grad} .
\end{aligned}
\end{equation}}

\section{Experiments}
\subsection{Datasets}
\textbf{TartanAir}\cite{wang2020tartanair} is a synthetic dataset for visual SLAM, covering various challenging indoor and outdoor scenarios. It encompasses over a thousand stereo videos, totaling approximately 306K stereo pairs, making it well-suited for training our temporal model. In this study, we employ this dataset for pre-training and conduct ablation experiments on it.

\noindent\textbf{Sceneflow}\cite{mayer2016large} is a synthetic dataset containing three sub-datasets covering indoor and outdoor scenes. This dataset includes short stereo video sequences (10 frames), with large motions between frames. We utilize the dataset for temporal training and compare our method with others.

\noindent\textbf{KITTI}\cite{menze2015object} is a real-world dataset for autonomous driving. It provides a leaderboard to evaluate the methods on a test set consisting of 200 images. Our work utilizes video sequences provided by KITTI for inference.

\noindent\textbf{ETH3D SLAM}\cite{schops2019bad} is a real-world dataset for SLAM in indoor environments. It provides stereo video sequences, as well as depth and pose ground truth. We evaluated the generalization performance of our method on this dataset.

\subsection{Implementation Details}
Our method is based on the implementation of RAFT-Stereo\cite{lipson2021raft}. For both training and testing, we set the number of refinement iterations $N$ to 5. During the training, we utilize a slice of stereo sequences as input and output the corresponding disparities frame by frame. We compute the loss for each frame's output and accumulate the gradients across all frames before updating the parameters. In our experiment, we set the sequence length to 2 or 4 frames for training. During the inference process, our method takes a video stream of arbitrary length as input and outputs disparity predictions in an online manner.
For the hyperparameters, we set \(\theta = 0.3\), \(\eta = 0.5\), \(\gamma = 0.9\), and \(\lambda_{\text{dc}}\) and \(\lambda_{\text{gdp}}\) to 0.1 and 1.2, respectively. We use the AdamW optimizer and a one-cycle learning rate schedule where the maximum learning rate is set to 0.0002. The batch size is set to 8 for all the experiments. For the TartanAir dataset, we use all data (including Easy and Hard) for the ablation study, and the detailed training-testing split can be found in the supplementary material. We train for 100k steps on this dataset. To make a fair comparison with TemporalStereo\cite{zhang2023temporalstereo}, we retrain our model with a sequence length of 2 and 200k steps according to their train-test split.
For the SceneFlow dataset, we train with the provided sequence data for 200k steps, with a sequence length of 2.
For the KITTI dataset, due to the lack of temporally annotated data, we follow the setting of TemporalStereo\cite{zhang2023temporalstereo} and train our model using pseudo-label data exported by the SOTA method LEAStereo\cite{cheng2020hierarchical} on the KITTI raw data. We train for 50k steps with a sequence length of 4 and a max learning rate of 0.0001, based on the model pre-trained on TartanAir. All pose data are either provided by the dataset or derived from SLAM algorithms. All experiments are conducted on two NVIDIA A40 GPUs.

\begin{table}[!t]
\centering
\scriptsize
\resizebox{\textwidth}{!}{
    \begin{tabular}{c|cc|cccc|ccc|cc|ccc|cc}
    \hline
    \multirow{3}{*}{Setting} &
      \multirow{3}{*}{\begin{tabular}[c]{@{}c@{}}Seq\\ length\end{tabular}} &
      \multirow{3}{*}{N} &
      \multirow{3}{*}{\begin{tabular}[c]{@{}c@{}}Past disp\\ \& state\end{tabular}} &
      \multirow{3}{*}{\begin{tabular}[c]{@{}c@{}}State \\ fusion\end{tabular}} &
      \multirow{3}{*}{\begin{tabular}[c]{@{}c@{}}TDC\\ module\end{tabular}} &
      \multirow{3}{*}{\begin{tabular}[c]{@{}c@{}}Dual-space\\ refinement\end{tabular}} &
      \multicolumn{5}{c|}{ALL} &
      \multicolumn{5}{c}{OCC} \\
      \cline{8-17} &
      & & & & & &
      \begin{tabular}[c]{@{}c@{}}>1px\\ (\%)\end{tabular} &
      \begin{tabular}[c]{@{}c@{}}>3px\\ (\%)\end{tabular} &
      \begin{tabular}[c]{@{}c@{}}EPE\\ (px)\end{tabular} &
      \begin{tabular}[c]{@{}c@{}}$|\Delta d|$\\ (px)\end{tabular} &
      \begin{tabular}[c]{@{}c@{}}${ Relu(\Delta e)}$\\ (px)\end{tabular} &
      \begin{tabular}[c]{@{}c@{}}>1px\\ (\%)\end{tabular} &
      \begin{tabular}[c]{@{}c@{}}>3px\\ (\%)\end{tabular} &
      \begin{tabular}[c]{@{}c@{}}EPE\\ (px)\end{tabular} & 
      \begin{tabular}[c]{@{}c@{}}$|\Delta d|$\\ (px)\end{tabular} &
      \begin{tabular}[c]{@{}c@{}}${ Relu(\Delta e)}$\\ (px)\end{tabular} \\
      \hline
    (A) & 2 & 5  &   &  &   &   & 8.04 & 2.98 & 0.53 & 0.35 & 0.16 & 34.91 & 16.60 & 1.63 & 1.15 & 0.47 \\
    (B) & 2 & 32 &   &  &   &   & 6.02 & 2.26 & 0.41 & 0.28 & 0.13 & 25.89 & 12.39 & 1.28 & 0.93 & 0.39 \\ \hline
    (C) & 2 & 5  & \checkmark &   &   &   & 8.77  & 3.27 & 0.60 & 0.29 & 0.12 & 35.04 & 16.01 & 1.67 & 0.94 & 0.34 \\
    (D) & 2 & 5  & \checkmark & \checkmark  &   &   & 6.98 & 2.55 & 0.45 & 0.25 & 0.10 & 29.89 & 13.33 & 1.32 & 0.86 & 0.31 \\
    (E) & 2 & 5  & \checkmark & \checkmark  & \checkmark &   & 6.10 & 2.33 & 0.42 & 0.23 & 0.10 & 25.97 & 12.40 & 1.21 & 0.78 & 0.28 \\
    (F) & 2 & 5  & \checkmark & \checkmark  &   & \checkmark & 6.28  & 2.42 & 0.42 & 0.21 & 0.09 & 25.89 & 11.86 & 1.18 & 0.74 & 0.28 \\
    (G) & 2 & 5  & \checkmark & \checkmark  & \checkmark & \checkmark & 5.98 & 2.19 & 0.40 & 0.21 & 0.09 & 24.67 & 11.28 & 1.14 & 0.71 & 0.27 \\ \hline
    (H) & 4 & 5  & \checkmark & \checkmark & \checkmark & \checkmark & \textbf{5.84} & \textbf{2.05} & \textbf{0.39} & \textbf{0.20} & \textbf{0.08} & \textbf{23.85} & \textbf{10.50} & \textbf{1.06} & \textbf{0.68} & \textbf{0.26} \\
    (I) & 4 & 5  &  & \checkmark & \checkmark & \checkmark & 6.50 & 2.29 & 0.43 & 0.28 & 0.13 & 26.80 & 12.48 & 1.26 & 0.90 & 0.38 \\
    \hline
    \end{tabular}}
    \caption{{\textbf{Ablation study on the TartanAir dataset.} ALL and OCC respectively represent metrics for all regions and occluded regions.}}
    \label{tab:ablation}
\end{table}
\subsection{Ablation Study and Analysis}
\noindent\textbf{Ablation Study}   
We demonstrate the effectiveness of our designs through ablation experiments conducted on the TartanAir dataset. As shown in Table \ref{tab:ablation}, we investigate the impacts of sequence length for training, temporal information (\ie Past disp \& state), temporal state fusion, temporal disparity completion (TDC) module, and dual-space refinement module on the accuracy in all regions and occluded regions. 
Initial settings (A) and (B) serve as baselines, representing RAFT-Stereo\cite{lipson2021raft} with 5 and 32 iterations respectively. For a fair comparison, we set the sequence length during training to 2 for them, even though they don't use temporal information.
Similar to XR-Stereo\cite{cheng2024stereo}, setting (C) utilizes the past disparity and hidden states as initialization, but direct iterations on this initialization reduce performance, as analyzed in Section \ref{sec:state fusion}.
As shown by the results of setting (D), the performance is improved in both all and occluded regions compared with (C), demonstrating the necessity of temporal state fusion.
The accuracy of (E) shows further improvement, particularly on the 1-px error rate, which benefits from the well-initialized disparity provided by the TDC module. This enables the refinement module to be devoted to improving the fine-grained matching.
(F) demonstrates that the dual-space refinement effectively improves the results in occluded regions. 
By combining (E) and (F), (G) achieves an even more precise stereo matching capability, exemplifying the synergy between the modules.
Furthermore, we extend the training sequence length from 2 to 4 and get a better disparity output, suggesting that the performance of our model scales with a more extensive temporal context.
Setting (I) corresponds to the model of (H) in the single-frame mode. Although there is a slight decrease in accuracy compared to setting (H), the results are still superior to (A). This demonstrates the versatility of our method in the single-frame and multi-frame mode, while also emphasizing the importance of temporal information.
To sum up, our TC-Stereo (G) and (H), with only 5 iterations, significantly improve the results of RAFT-Stereo (A) and even surpass the performance of RAFT-Stereo with 32 iterations (B) on all metrics, greatly enhancing efficiency.

\noindent\textbf{Temporal Consistency}  We design two types of metrics to evaluate the temporal consistency. For the first one, We convert the predicted disparity map $d^{t+1}$ to the image coordinate at time $t$ through poses and ground truth optical flow, denoting as $\widetilde{d}^{t+1}$.
We use the absolute difference between $\widetilde{d}^{t+1}$ and $d^t$, denoted as $|\Delta d|=|\widetilde{d}^{t+1}-d^t|$, to evaluate temporal consistency.
However, evaluating temporal absolute differences alone is one-sided. If a disparity \( d^t \) is incorrect, the model should be allowed to make correct predictions at the next time step \( t+1 \), even if this results in temporary inconsistencies. 
To evaluate temporal consistency comprehensively, we design a more lenient metric to allow the temporal change towards \(d^{gt}\).
Concretely, we calculate the error maps \(e=|d-d^{gt}|\) for timestamp $t$ and $t+1$, and then compute the change in error, \(\Delta e = \widetilde{e}^{t+1}-e^t\). 
We use \(Relu(\Delta e)\) to evaluate the extent of error divergence over time (\ie, error increased from the previous frame for the same 3D point).
Table \ref{tab:ablation} proves that our final models, (G) and (H), achieve the best temporal consistency and convergence compared to other settings. 
Settings (D), (E), and (F) respectively demonstrate the effectiveness of the state fusion, TDC module, and Dual-space refinement module in improving temporal consistency.
Notably, (G) is comparable in accuracy to the baseline with 32 iterations (B), yet (G) significantly outperforms (B) in temporal metrics, especially in occlusion areas, further demonstrating the effectiveness of our TC-Stereo in mitigating temporal inconsistencies. Additionally, although (I) surpasses (D) in accuracy metrics, it is inferior to (D) in temporal metrics, further revealing the crucial role of temporal information in improving temporal consistency. We also provide the results of the 3-pixel error of consistency metrics in the supplementary materials.

\begin{figure}[!t]
\centering
\begin{subfigure}[b]{0.49\textwidth}
    \includegraphics[width=\textwidth]{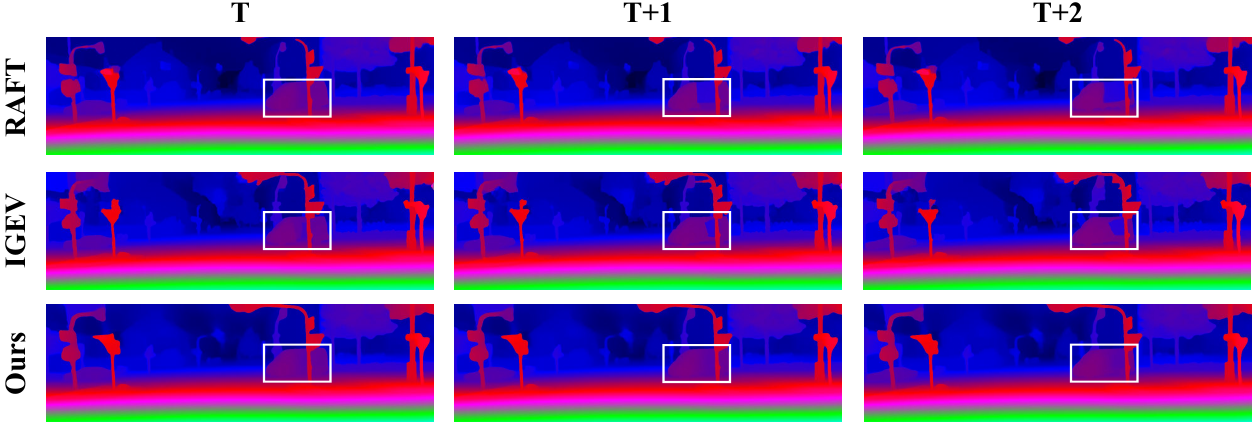}
    \caption{}
    \label{fig:kitti_a}
\end{subfigure}
\hfill
\begin{subfigure}[b]{0.47\textwidth}
    \includegraphics[width=\textwidth]{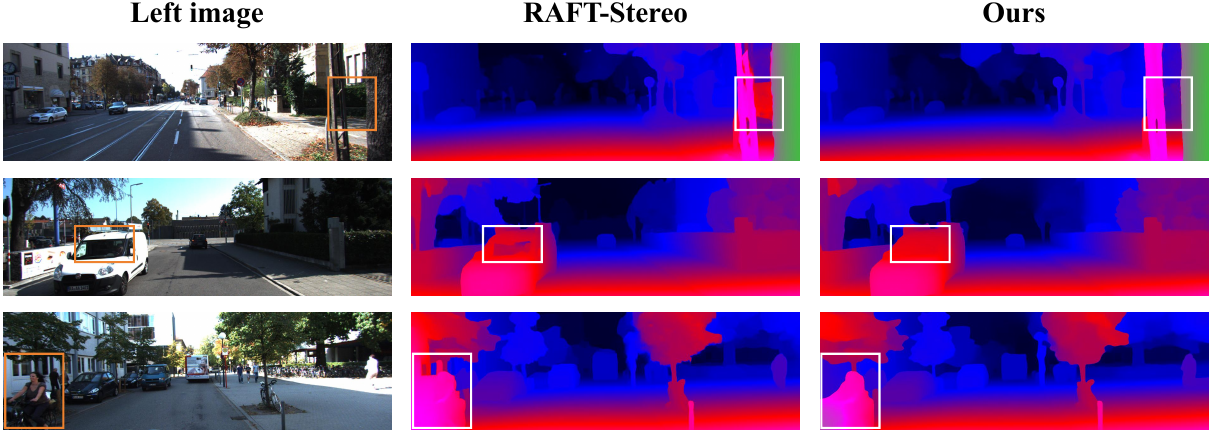}
    \caption{}
    \label{fig:kitti_b}
\end{subfigure}
\label{fig:kitti_vis}
\caption{
     \textbf{Visualizations on KITTI 2015.} (a) Comparison of temporal disparity sequences from RAFT-Stereo\cite{lipson2021raft}, IGEV\cite{xu2023iterative} and our method. (b) Comparison of disparities in ill-posed regions between RAFT-Stereo\cite{lipson2021raft} and our method.}
\end{figure}

We present visualizations of disparity sequences to illustrate the improvement in temporal consistency achieved by our method. As shown in Fig. \ref{fig:kitti_a}, our method produces more consistent disparity maps in the highlighted white box areas compared to other SOTA methods. More visualizations of temporal consistency are given in the appendix and our project website.

\begin{table}[!b]
\centering
\resizebox{0.98\textwidth}{!}{
\begin{tabular}{cccccccc}
\hline
\textbf{Method} &  RAFT-Stereo\cite{lipson2021raft} & LEAStereo\cite{cheng2020hierarchical} &  DLNR\cite{zhao2023high} & DynamicStereo\cite{karaev2023dynamicstereo} &  TemporalStereo\cite{zhang2023temporalstereo} &  CODD\cite{li2023temporally} & Ours \\ \hline
\textbf{EPE (px)}    & 0.65            &  0.78         & \textbf{0.47}     &        0.71 &  0.53   & 0.60 & 0.50     \\
\textbf{>3px (\%)}    & 3.13            & -          &     -         &      3.38   &  2.75         & 2.80 &\textbf{2.39}     \\ 
\textbf{>1px (\%)}    & 6.67            & 7.82         &   5.38       &      7.57   &  6.38          & - & \textbf{4.93}     \\ \hline
\end{tabular}
}
\caption{Benchmark results on Sceneflow dataset.}
\label{tab:sceneflow}
\end{table}

\begin{figure}[!t]
\centering
\begin{minipage}[!t]{0.49\textwidth}
    \centering
    \includegraphics[width=\textwidth]
    {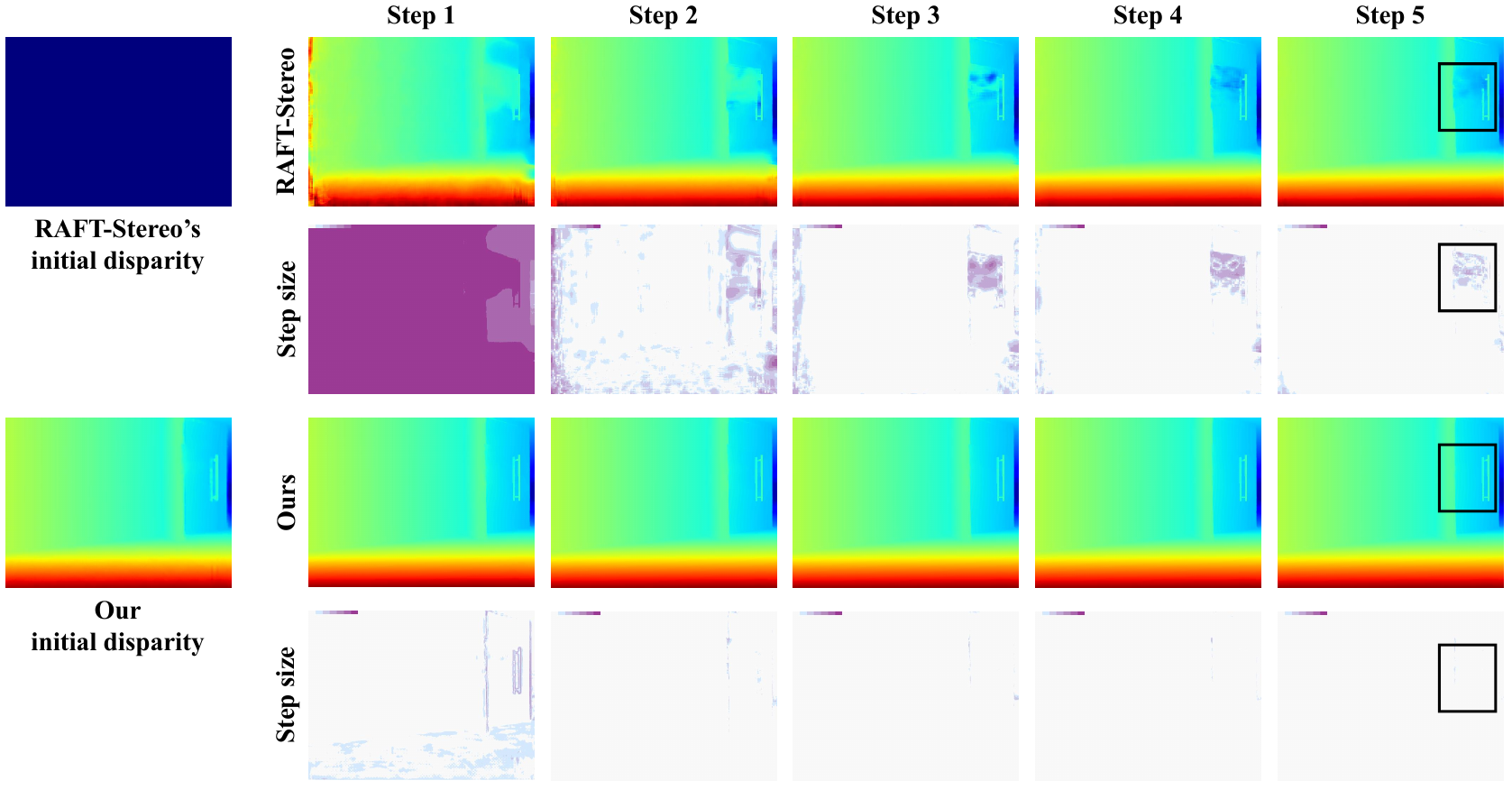}
    \caption{
     Visualizations of the disparity map and the update step size at each iteration. }
    \label{fig:step size}
\end{minipage}
  \hfill
\begin{minipage}[!t]{0.5\textwidth}
\centering
\resizebox{0.93\textwidth}{!}{
\begin{tabular}{c|ccc|ccc|c}
\hline
\multirow{3}{*}{Method} &
  \multicolumn{3}{c|}{Noc} &
  \multicolumn{3}{c|}{All} &
  \multirow{3}{*}{\begin{tabular}[c]{@{}c@{}}Time\\ (s)\end{tabular}} \\ \cline{2-7}
 &
  \begin{tabular}[c]{@{}c@{}}D1-bg\\ (\%)\end{tabular} &
  \begin{tabular}[c]{@{}c@{}}D1-fg\\ (\%)\end{tabular} &
  \begin{tabular}[c]{@{}c@{}}D1-all\\ (\%)\end{tabular} &
  \begin{tabular}[c]{@{}c@{}}D1-bg\\ (\%)\end{tabular} &
  \begin{tabular}[c]{@{}c@{}}D1-fg\\ (\%)\end{tabular} &
  \begin{tabular}[c]{@{}c@{}}D1-all\\ (\%)\end{tabular} &
   \\ \hline
RAFT-Stereo \cite{lipson2021raft}    & 1.45          & 2.94          & 1.69          & 1.58          & 3.05          & 1.82          & 0.38          \\
CREStereo \cite{li2022practical}      & 1.33          & 2.6           & 1.54          & 1.45          & 2.86          & 1.69          & 0.41          \\
LEAStereo \cite{cheng2020hierarchical}     & 1.29          & 2.65          & 1.51          & 1.4           & 2.91          & 1.65          & 0.3           \\
DLNR \cite{zhao2023high}          & 1.45          & 2.39 & 1.61          & 1.6           & 2.59 & 1.76          & 0.3           \\
IGEV-Stereo \cite{xu2023iterative}   & 1.27          & 2.62          & 1.49          & 1.38          & 2.67          & 1.59          & 0.18          \\
PCW-Net \cite{shen2022pcw}       & 1.26          & 2.93          & 1.53          & 1.37          & 3.16          & 1.67          & 0.44          \\ \hline
HITNet \cite{tankovich2021hitnet}         & 1.54          & 2.72          & 1.74          & 1.74          & 3.2           & 1.98          & \textbf{0.02} \\
TemporalStereo \cite{zhang2023temporalstereo} & 1.52          & 2.58          & 1.70           & 1.61          & 2.78          & 1.81          & 0.04          \\ \hline
TC-Stereo(ours)     & \textbf{1.21} & \textbf{2.24}           & \textbf{1.38} & \textbf{1.29} & \textbf{2.33}          & \textbf{1.46} & 0.09          \\ \hline
\end{tabular}
}
\captionof{table}{ Benchmark results of KITTI2015.}
\label{tab:kitti}
\end{minipage}
\end{figure}

\noindent\textbf{Local Searching Range and Fast Convergence} 
Fig. \ref{fig:step size} illustrates the disparity maps and update step sizes of our method and RAFT-Stereo, step by step. It can be seen that our method iterates based on a well-initialized disparity, providing a local disparity searching range. This leads to smaller update step sizes and faster convergence to the ground truth.

\noindent\textbf{Improvement on Ill-posed Areas} Fig. \ref{fig:kitti_b} shows the results of occlusion, transparent areas, and out-of-view areas, respectively. Thanks to the temporal information and the design of dual-space refinement, our method achieved smoother disparity output and clearer boundaries in these ill-posed regions.

\noindent\textbf{Robustness Analysis}  Eq. \ref{projection} is based on the static scene assumption, resulting in unreliable initial disparities in dynamic areas. However, our method remains robust in ordinary dynamic scenes since we only use the temporal information for initialization. Disparities in dynamic regions will be corrected by iterative refinements. The moving car in Fig. \ref{fig:kitti_a} is a good example of this robustness. Additionally, we discuss the robustness of our method on incorrect poses and large motions in the supplementary materials.

\begin{figure}[!t]
\centering
\includegraphics[width=0.7\linewidth]{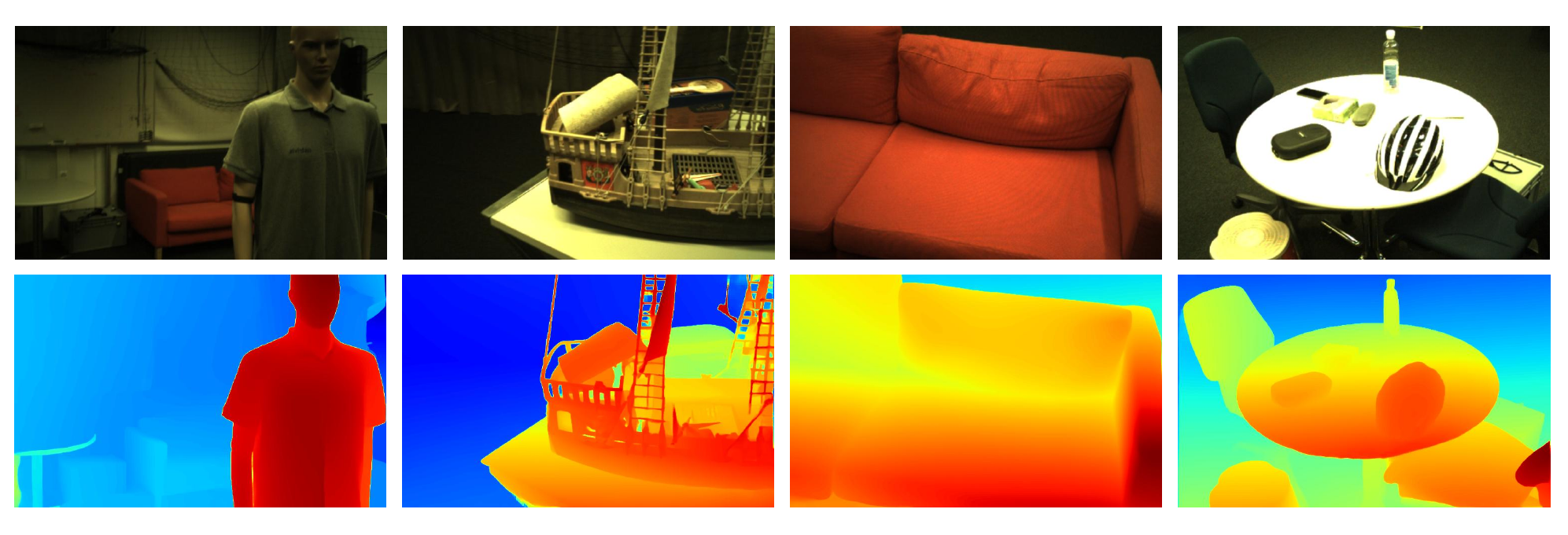}
\caption{ Zero-shot qualitative results of our method on ETH3D SLAM dataset.}
\label{fig:generalization}
\end{figure}

\noindent \textbf{Zero-shot Generalization} We evaluated the zero-shot generalization performance on the ETH3D SLAM dataset, as shown in Table \ref{tab: tartan and eth} (right). All models are only trained on SceneFlow. The metrics exclude regions with EPE $\!\!>\!\!50$ due to significant noise of depth ground truth around object boundaries. Our method surpasses other methods in generalization with fewer iterations. We also provide the qualitative results of our method on ETH3D SLAM, as shown in Fig. \ref{fig:generalization}.

\begin{table}[!t]
\begin{subtable}[b]{0.49\textwidth}
\centering
\resizebox{\textwidth}{!}{
\begin{tabular}{c|ccc|ccc}
\hline
\multirow{2}{*}{Method} & \multicolumn{3}{c|}{OCC} & \multicolumn{3}{c}{ALL} \\ \cline{2-7} 
 &
  \multicolumn{1}{c}{>1px} &
  \multicolumn{1}{c}{>3px} &
  \multicolumn{1}{c|}{EPE} &
  \multicolumn{1}{c}{>1px} &
  \multicolumn{1}{c}{>3px} &
  \multicolumn{1}{c}{EPE} \\ \hline
TemporalStereo*\cite{zhang2023temporalstereo}          & 39.78       & 18.46       & 1.60       & 11.18  & 3.81   & 0.66  \\
RAFT-Stereo\cite{lipson2021raft} (N=32) & \textbf{34.20}     & 16.67       &1.45     & \textbf{8.25} & 3.33  & 0.53  \\
Ours (N=5)                    & 34.68       & \textbf{16.36}   &  \textbf{1.36}      & 8.51   & \textbf{3.31} &\textbf{0.52}     \\ \hline
\end{tabular}
}
\end{subtable}
\hfill
\begin{subtable}[b]{0.49\textwidth}
\centering
\scalebox{.65}[.65]{%
\begin{tabular}{c|ccc}
\hline
\multirow{1}{*}{Method} & \begin{tabular}[c]{@{}c@{}}$>\!\!1$px\\ (\%)\end{tabular} & \begin{tabular}[c]{@{}c@{}}$>\!\!3$px\\ (\%)\end{tabular} & \begin{tabular}[c]{@{}c@{}}EPE\\ (px)\end{tabular}  \\ \hline
IGEV-Stereo\cite{xu2023iterative} ($N=32$) & 37.72     & 24.82           & 6.41                \\
RAFT-Stereo\cite{lipson2021raft} ($N=32$) & 38.19     & 24.95            & 6.36                \\
Ours ($N=5$)         & \textbf{37.00}     & \textbf{23.29}            & \textbf{6.27}         \\ \hline
\end{tabular}
}
\end{subtable}
\caption{ \textbf{Left:} Results of Tartanair dataset based on the train-test splits of TemporalStereo\cite{zhang2023temporalstereo}. * represents using the weights provided by the author. \textbf{Right:} Zero-shot results on ETH3D SLAM dataset. }
\label{tab: tartan and eth}
\end{table}

\subsection{Benchmark Results}

\noindent \textbf{Sceneflow}  We conducted evaluations on the SceneFlow dataset, as shown in Table \ref{tab:sceneflow}. Overall, our method outperforms the other methods except DLNR on the EPE metric, which may be attributed to their use of a more powerful backbone and more iterations. Our method exhibits a clear advantage over others in the $>\!\!1$px and $>\!\!3$px error rate metrics.

\noindent \textbf{TartanAir}   We compared our method with TemporalStereo\cite{zhang2023temporalstereo} and RAFT-Stereo\cite{lipson2021raft} on the TartanAir dataset using the train-test split of TemporalStereo. As shown in Table \ref{tab: tartan and eth} (left), our model surpassed RAFT-Stereo with only 5 iterations. Furthermore, it demonstrates significant advantages over the other two methods in terms of the EPE metric in occluded regions. This indicates that our design can enhance performance in ill-posed areas.

\noindent \textbf{KITTI 2015}  We present a comprehensive comparison of our method with SOTA methods on the KITTI 2015 dataset. As shown in Table \ref{tab:kitti}, our proposed method, TC-Stereo, outperforms the other methods in most key metrics. 
Specifically, it achieves the lowest error rates in both non-occluded (Noc) and all regions, with a D1-all error rate of 1.46\%, which is an 8.18\% improvement over the best-performing IGEV-Stereo\cite{xu2023iterative}. 
Among all these methods, our method exhibits the smallest difference between the Noc and All metrics, demonstrating our superior performance in occluded areas. 
Additionally, TC-Stereo showcases exceptional efficiency, with a processing time of only 0.09 seconds per frame, which is significantly faster than other methods with high performance. 
This performance shows that TC-Stereo excels in both accuracy and speed, making it well-suited for online applications.

\section{Limitation}
Despite the robustness of our model for ordinary dynamic regions and pose noise, it remains sensitive to large motions of dynamic objects, large errors in camera pose, and little overlap between adjacent frames, as we discuss in the supplementary materials. In these challenging cases, our method degrades to searching for the ground truth in a global disparity range like RAFT-Stereo. 

\section{Conclusion}
In this paper, we proposed a temporally consistent stereo matching method that exploits temporal information through our temporal disparity completion module and temporal state fusion module, providing a well-initialized disparity map and hidden states.
We further proposed a dual-space refinement module that iteratively refines the temporally initialized results in both disparity and disparity gradient spaces, improving estimations in ill-posed regions.
Experiments demonstrated that our method effectively mitigates temporal inconsistency in video stereo matching and achieves SOTA results on challenging datasets.

\section*{Acknowledgements}
This work was supported by the Natural Science Foundation of Shenzhen under Grant No. JCYJ20230807142703006, Natural Science Foundation of China (NSFC) under Grants No. 62172041 and No. 62176021, and Key Research Platforms and Projects of the Guangdong Provincial Department of Education under Grant No.2023ZDZX1034.

%
%
\bibliographystyle{splncs04}
\bibliography{main}
\end{document}